\documentclass[10pt,twocolumn,letterpaper]{article}

\usepackage{iccv}
\usepackage{times}
\usepackage{epsfig}
\usepackage{graphicx}
\usepackage{amsmath}
\usepackage{amssymb}
\usepackage{algorithm}
\usepackage{algpseudocode}
\usepackage{multirow}

\usepackage[pagebackref=true,breaklinks=true,letterpaper=true,colorlinks,bookmarks=false]{hyperref}

\iccvfinalcopy 

\def\iccvPaperID{3570} 

\begin{document}

\title{A Study on Action Detection in the Wild}

\author{Yubo Zhang, Pavel Tokmakov, Martial Hebert\\
Carnegie Mellon University\\
{\tt\small yuboz,ptokmako,hebert@andrew.cmu.edu}
\and
Cordelia Schmid\\
Google Research\\
{\tt\small cordelias@google.com}
}

\maketitle

\begin{abstract}
The recent introduction of the AVA dataset for action detection has caused a renewed interest to this problem. Several approaches have been recently proposed that improved the performance. However, all of them have ignored the main difficulty of the AVA dataset - its realistic distribution of training and test examples. This dataset was collected by exhaustive annotation of human action in uncurated videos. As a result, the most common categories, such as `stand' or `sit', contain tens of thousands of examples, whereas rare ones have only dozens. In this work we study the problem of action detection in a highly-imbalanced dataset. Differently from previous work on handling long-tail category distributions, we begin by analyzing the imbalance in the test set. We demonstrate that the standard AP metric is not informative for the categories in the tail, and propose an alternative one - Sampled AP. Armed with this new measure, we study the problem of transferring representations from the data-rich head to the rare tail categories and propose a simple but effective approach.
\end{abstract}


\section{Introduction}

Being able to recognize every concept in the world is one of the ultimate goals of computer vision. For tasks such as image classification \cite{he2015delving,krizhevsky2012imagenet}, action recognition \cite{ji20133d,simonyan2014two}, or semantic segmentation~\cite{everingham2010pascal}, the performance on many benchmark datasets has practically saturated \cite{han2017deep,stroud2018d3d}. However, these tasks can only be considered as solved in the closed world of these datasets, where often the distribution of examples per category is artificially balanced both in the training and test sets. In the real world, this distribution is highly imbalanced, with a few categories covering most of the data (so called head of the distribution), and the rest having only several examples per category (so called tail). As a result, methods developed using standard benchmark datasets end up being not well suited for the real world. 

\begin{figure}
\begin{center}
\includegraphics[width=1.0\linewidth]{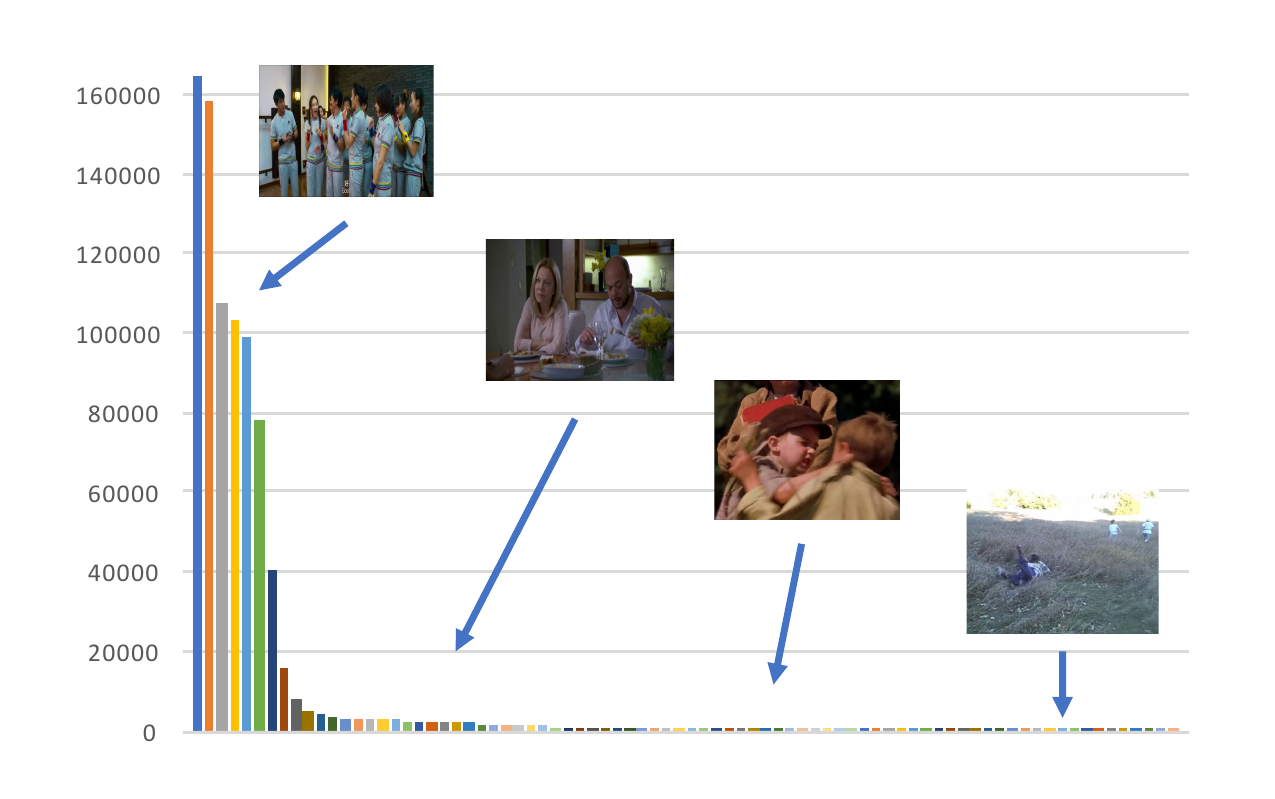}
\end{center}
   \caption{The distribution of the action categories in the training set of AVA~\cite{gu2017ava}. This dataset is collected by exhaustively annotating real world videos and exhibits a strong imbalance in the number of examples between the common and rare classes. As a results, state-of-the art approaches achieve very low  performance on the categories in the tail.}
\label{fig:distribution}
\end{figure}

This problem is especially severe in the action recognition domain, where many actions are extremely rare. Consider the recent AVA dataset~\cite{gu2017ava} for action detection. To collect this dataset, the authors have extensively annotated all the human actions that appear in the 107.5 hours of movie footage, resulting in a realistic distribution of training and test examples, see Figure~\ref{fig:distribution}. In particular, the most frequent category {\tt stand} has 164932 training examples, whereas the rarest category {\tt extract} has only 7. State-of-the-art methods on this dataset~\cite{feichtenhofer2018slowfast,girdhar2018video} achieve very low performance on most of the tail categories.

\begin{figure*}
\begin{center}
\includegraphics[width=1.0\linewidth]{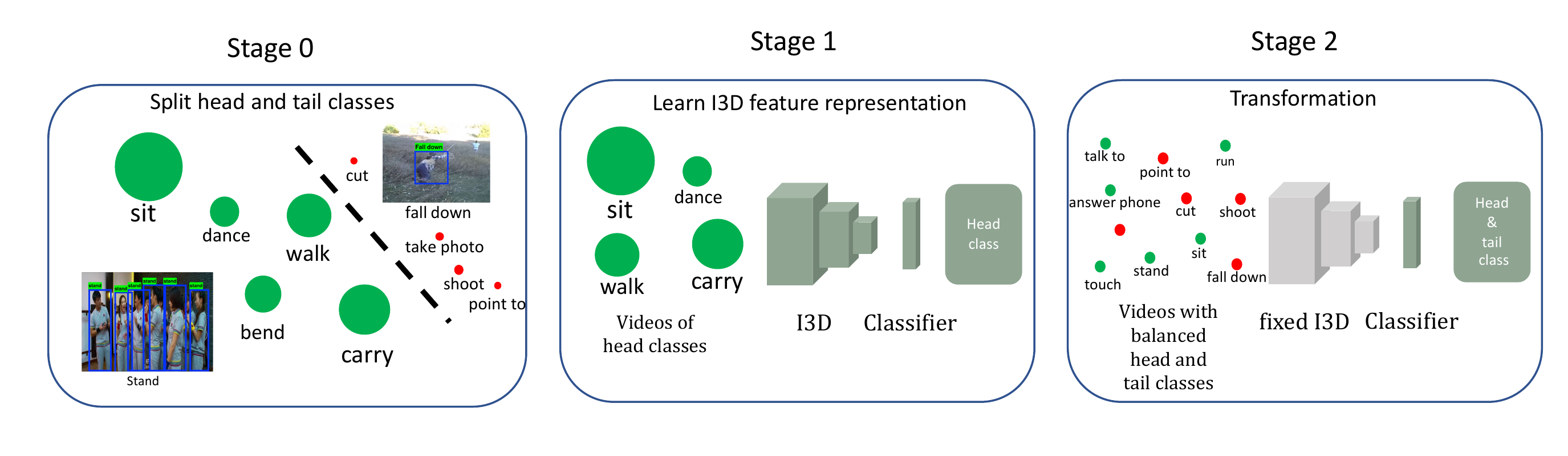}
\end{center}

   \caption{Our proposed training schema. We first split the categories into head and tail on Stage 0. The size of the circle indicates the number of samples belonging to class. In stage one, we train an I3D feature representation with only head categories. In stage two, we rebalance the samples of head and tail classes to fine-tune the classification module.}
\label{fig:model}
\end{figure*}

Handling long-tail distributions in the training set has been studied in the image domain. The most common strategy is rebalancing the training set~\cite{Japkowicz:2002:CIP:1293951.1293954,Ling:1998:DMD:3000292.3000304,zhong2016towards}. We demonstrate that naively applying this approach to highly imbalanced video datasets leads to a decrease in performance. We then analyze the reasons for this phenomenon and propose an effective alternative. In~\cite{wang2017learning} the authors learn to transfer information from head to tail categories via meta-learning. Their approach, however, is based on an assumption that a large set of head categories is available for meta-training, which does not hold for AVA. We also propose to transfer information from head to tail categories, but our method does not have such assumptions, is simpler, and more efficient. Finally, Zhao et al.~\cite{zhao2017open} utilize the WordNet hierarchy~\cite{miller1995wordnet} to aid in transferring information between semantically related categories. While promising, this approach is not directly applicable to the domain of actions. Notice that none of these works address the imbalance in the distribution of \textit{test} examples.

In this paper we study the problem of action detection in the wild, using the AVA dataset as the motivating example. In contrast to previous work, we start by analyzing the data imbalance problem in the test distribution. We demonstrate that the standard AP metric used for evaluating action detection is not informative for the tail classes: the overwhelming majority of the test examples are negatives for these categories, thus the false positive rate dominates the score. Based on this observation we proposed an alternative measure: sampled AP. It is computed by taking multiple balanced samples from the test set and averaging the score over the samples. With a sufficient number of samples, this measure preserves the standard AP performance for the head categories, while addressing the aforementioned issue for the categories in the tail. Note that AP often does not allow to make conclusion about performance of a model on the tail categories, whereas SAP is an \textit{actionable metric} for learning to recognize rare classes.

Armed with the new measure, we analyze the problem of detecting the actions in the tail. Firstly, we experiment with naive rebalancing of the training examples. This approach, however, results in a decrease in performance for both head and tail categories. Recall that in AVA the most frequent category has 22560 times more examples than the least frequent one, thus the rebalanced dataset mostly consists of duplicated examples from the tail, not allowing the model to learn a high quality representation within a realistic time budget. To mitigate this issue, we propose to explicitly split the dataset into the head and tail parts and only train our initial model on the data-rich head categories. We then freeze the model trunk, and finetune the classifier using the whole dataset with balancing, see Figure~\ref{fig:model}. This simple approach improves the performance of the tail categories by a significant margin. We also study several other variants of model finetuning, and demonstrate that they lead to an inferior performance. 

To summarize, this work makes two main contributions:

\begin{enumerate}
    
    \item We propose a novel metric for action detection: sampled mAP. It preserves the properties of mAP for the categories in the head, while allowing to better analyze the tail categories.
    
    \item We propose a simple, yet efficient, approach which significantly improves the performance of the categories in the tail. 
\end{enumerate}

\section{Related work}


\textbf{Long tail distributions} have been well studied in the image domain. Oversampling \cite{Ling:1998:DMD:3000292.3000304} and undersampling \cite{Japkowicz:2002:CIP:1293951.1293954} are two most common approaches in the literature. Oversampling rebalances the training set by duplicating the samples of the tail classes, whereas undersampling ignores some of the examples in the head classes. We demonstrate that naively applying such approaches to highly-imbalanced video datasets leads to a decrease in performance. Wang et al.~\cite{wang2017learning} proposed to use meta-learning to transfer information from head to tail categories. In particular, they learn a model that takes a classifier trained from a few examples and transforms into a large sample classifier. Their method, however, assumes that many large-sample categories are available in the meta-training stage. Our approach is not limited by such assumptions, and is also simpler and more efficient. Recently, Cui et al.~\cite{Cui2018Fine} proposed a two-stage method for transfer learning on fine-grained classification tasks, where the model is first trained on the whole dataset and then fine-tuned on a balanced subset. In contrast, we demonstrate that explicitly separating the training set into head and tail categories and only training the model on the head, as well as freezing the model weights in the balanced training stage leads to better results. 
With the development of deep learning methods, new kinds of loss functions \cite{Wang2016TrainingDN} have been proposed to mitigate data imbalance. In~\cite{zhao2017open} the authors propose to utilize the information about semantic similarity between categories to aid in transferring information between them. While this approach is promising, it can not be extended to the domain of actions in a straight-forward way. In object detection Lin et al. \cite{Lin2017FocalLF} proposed focal loss to down-weight the gradients of well-classified examples. It is, however, designed for relatively well balanced datasets, such as COCO. Our preliminary experiments on human pose categories in AVA have shown that focal loss in not effective in this more realistic scenario. Differently from all the methods above, our study focuses on significantly more imbalanced datasets, such as the AVA dataset for action detection, and addresses data imbalance not only in the training, but also in the test set.

\textbf{Action classification} assigns category labels to videos. Hand-crafted features \cite{wang2013dense} are, for example, obtained based on features generated by tracking pixels and aggregating information along the temporal axis. These pre-deep-learning methods have recently been outperformed by deep learning based models. An initial approach is based on two-stream networks~\cite{peng2016multi}, which separately process image and optical flow with two CNNs and merges the output. However, the capacity of these models is limited by only relying on 2D information. This limitation was addressed by~\cite{tran2015learning} who extended 2D CNN filters with an additional dimension, enabling learning of spatio-temporal features. Carreira and Zisserman~\cite{carreira2017quo} further extended this work and introduced Inflated 3D ConvNet (I3D) by integrating 3D filters into the state-of-the-art 2D architecture and bootstrapping the 3D filters from 2D filters pretrained on ImageNet. Wang et al.~\cite{wang2017non} further improved the performance by proposing non-local blocks that integrate information from distant locations in space and time. In this paper, we use the non-local I3D model to learn a spatio-temporal feature representation. Notice that the high representational power of these models comes at a price: they can easily overfit to the few training examples of the tail categories. We address this issue by proposing a new training scheme for transferring representations from head to tail classes.

\textbf{Action detection} is the task of spatially localizing the actors with bounding boxes and recognizing their actions. Early action detection methods~\cite{klaser2010human,prest2013explicit} extracted hand crafted features from videos and trained SVM classifiers. Early deep-learning based action detection models~\cite{Gkioxari_2015_CVPR,peng2016multi,saha2016deep,singh2017online,Weinzaepfel_2015_ICCV} are developed on top of 2-D object detection frameworks, where 2-D appearance and optical flow features are used for action classification. As a further step, Kalogeiton et al.~\cite{kalogeiton2017action} propose to take multiple frames as input and predict and classify short tubelets instead of single bounding boxes. Recently, many works propose to take videos as input and learn spatio-temporal features with 3D convolutional neural networks. TCNN~\cite{hou2017tube} uses C3D to extract features and results in a performance increase. Gu et al.~\cite{gu2017ava} rely on I3D features and demonstrate the benefit of taking longer video sequences as input. Recently, several works have increased the performance of these approaches. For instance, Feichtenhofer et al.~\cite{feichtenhofer2018slowfast} propose a slowfast network with two pathways, where one processes a video with a high FPS and the other one with a low FPS. Their model is able to capture long temporal dependencies as well as informative context improving the action detection performance. Girdhar et al.~\cite{girdhar2018video} further propose to use a transformer-style architecture to integrate relevant information temporally and spatially. Very recently, Zhang et al.~\cite{zhang2018structured} augment I3Ds with a structured module based on graph convolutions. 




\section{Background}

\begin{figure*}
\begin{center}
\includegraphics[width=1.0\linewidth]{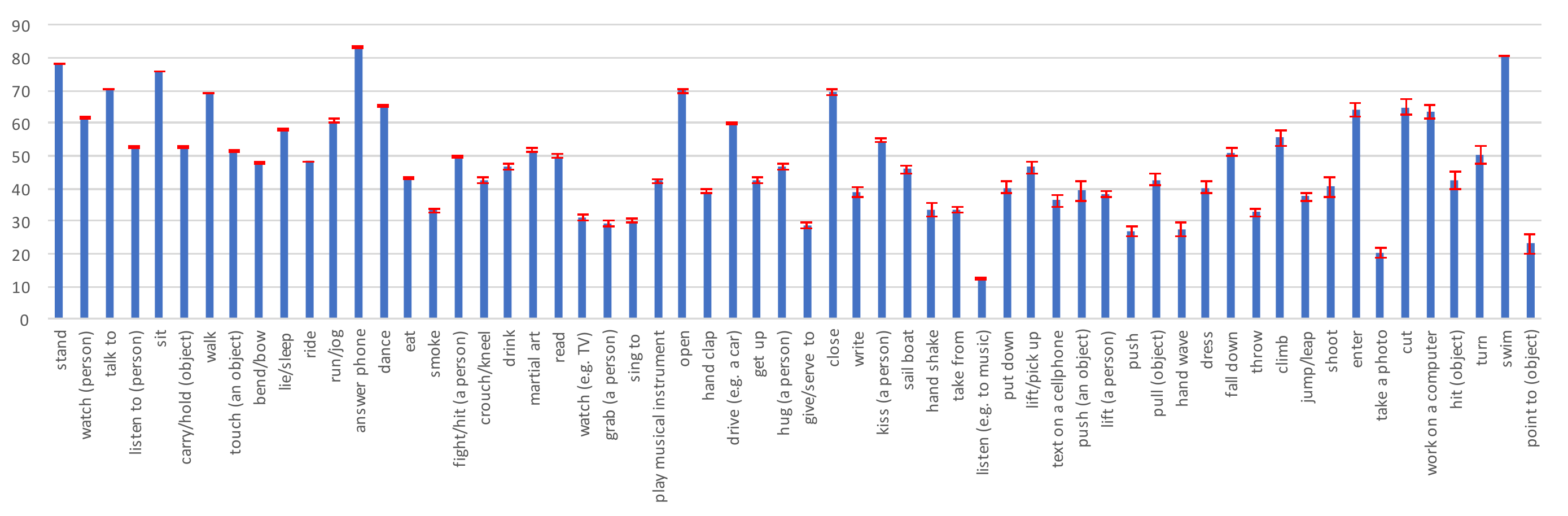}
\end{center}
   \caption{Visualization of the mean and standard deviation of our proposed SAP metric on our baseline approach. As shown in the figure, the values of standard deviation of all classes are small which proves that our proposed SAP metric is informative as well as practical.}
\label{fig:std_mean_metric}
\end{figure*}

We begin by introducing the dataset and models used in our study.

\subsection{Dataset and metric}

We perform our study using the AVA dataset~\cite{gu2017ava}. It consists of hours of raw videos collected from movies and exhaustively annotated with 80 human action categories including human pose, human-object manipulation and human-human interaction. The dataset is split into 211k training and 57k validation clips, each of which typically contains several action instances. The exhaustive labelling of all actions of all persons in all key frames at 1 Hz results in a Zipf's law type of imbalance across action categories as shown in Figure~\ref{fig:distribution}. In particular, a common action, like {\tt standing}, has 160k training and 43k test examples, whereas a rare action, like {\tt point to (an object)} only contains 96 training and 32 test instances. Such a distribution of the action categories is desirable, as it represent realistic scenarios. Yet, it makes both training and evaluating the model challenging. 

We use frame-based mean average precision(mAP) with intersection-over-union (IOU) threshold 0.5 for evaluation. Following the protocol in~\cite{gu2017ava}, we only evaluate on 60 categories which have more than 25 examples in the validation set. In addition to the standard AP metric, we also show results on our proposed sampled AP which, as discussed in Section~\ref{sec:measure}, is more informative for the categories in the tail. 

\subsection{Action detection model}

We study the action detection task, where actors are localized with bounding boxes and actions are recognized based on a spatio-temporal descriptor. 
Our action detection model takes three second video clips as input and uses the non-local Inflated 3D ConvNet (I3D) to extract a features. A state-of-the-art Mask-RCNN person detector~\cite{he2017mask} localizes actors in the middle frame of each three second video clip. We use ROI pooling to extract features for each bounding box and pass the resulting representations though a single fully connected layer to predict category scores. 


\subsection{Implementation details}

Our method is built on top of the Caffe2 framework. We use 2D ResNet-50 architecture and pretrain it on the ImageNet dataset~\cite{krizhevsky2012imagenet}. It is then inflated into 3D ConvNet and fine-tuned on the Kinetics dataset~\cite{carreira2017quo}. Our action detection model takes as input 36 frames from a 3 second video clip with 12 fps. The frames are first scaled to 272 $\times$ 272, an then randomly cropped to 256 $\times$ 256.

We train our model on a 8-GPU machine. For training the I3D backbone we use 3 video clips per GPU as a mini-batch with a total batch size of 24. For training the linear classifier in the transfer learning experiments, we use a batch size of 1000. We freeze parameters of batch normalization layers during training, and apply a drop out before the final layer with rate 0.3. We train the model using SGD with a learning rate of 0.00125 for 90K iteration, and then drop the learning rate by a factor of 10, and continue training for another 10K iterations. In the transfer learning stage, we train the classification layer for 1 epoch while linearly decaying learning rate from 0.001 to 0.0001.

\begin{figure}
\begin{center}
\includegraphics[width=1.0\linewidth]{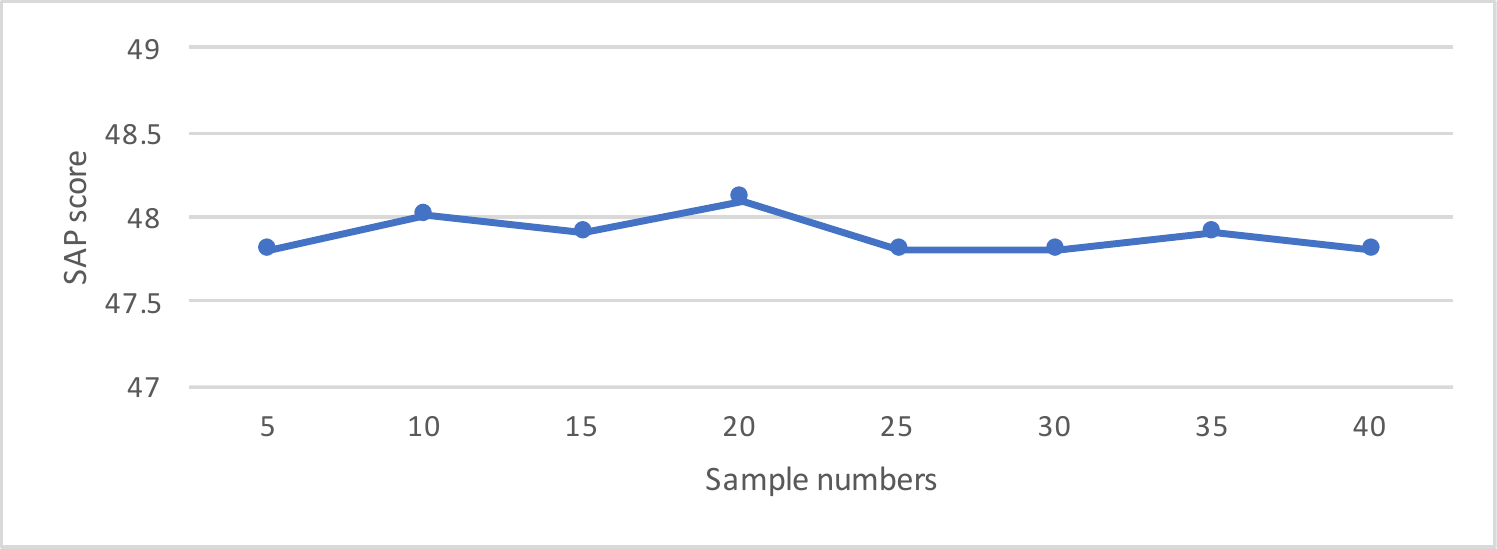}
\includegraphics[width=1.0\linewidth]{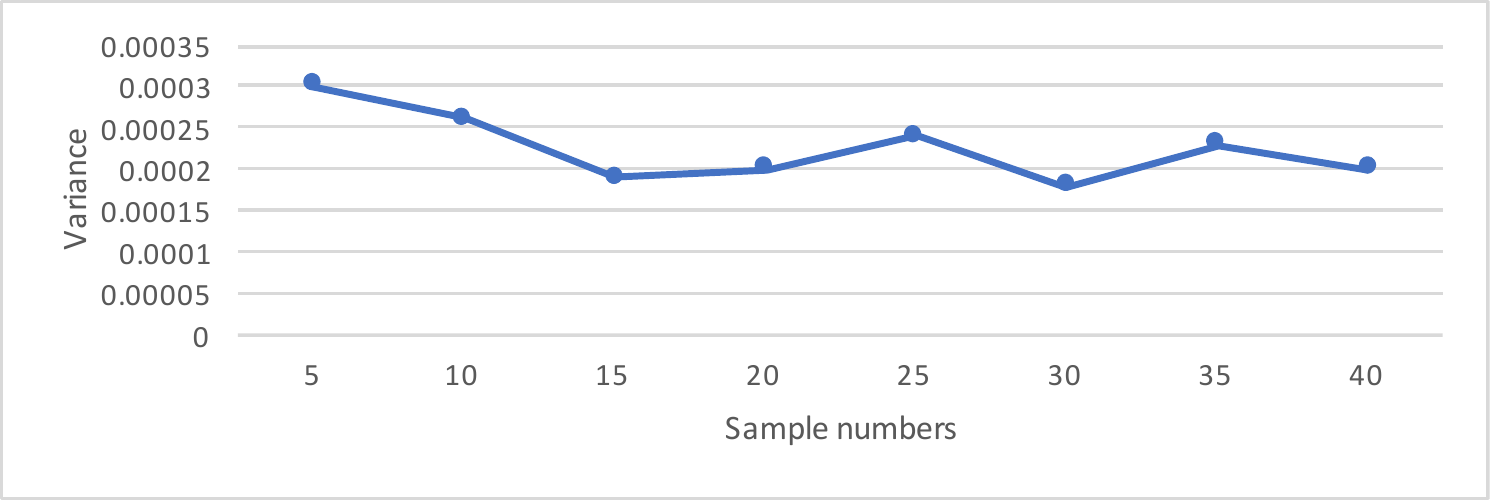}
\end{center}
   \caption{The mSAP score (top) and variance (bottom) with the number of samples varying from 5 to 40 for all test categories. As can be seen from the figures, our proposed metric exhibits low variance even with a few samples .}
\label{fig:metric_sample_number}
\end{figure}

\section{Measuring the imbalanced world}
\label{sec:measure}

\begin{figure*}
\begin{center}
\includegraphics[width=1.0\linewidth]{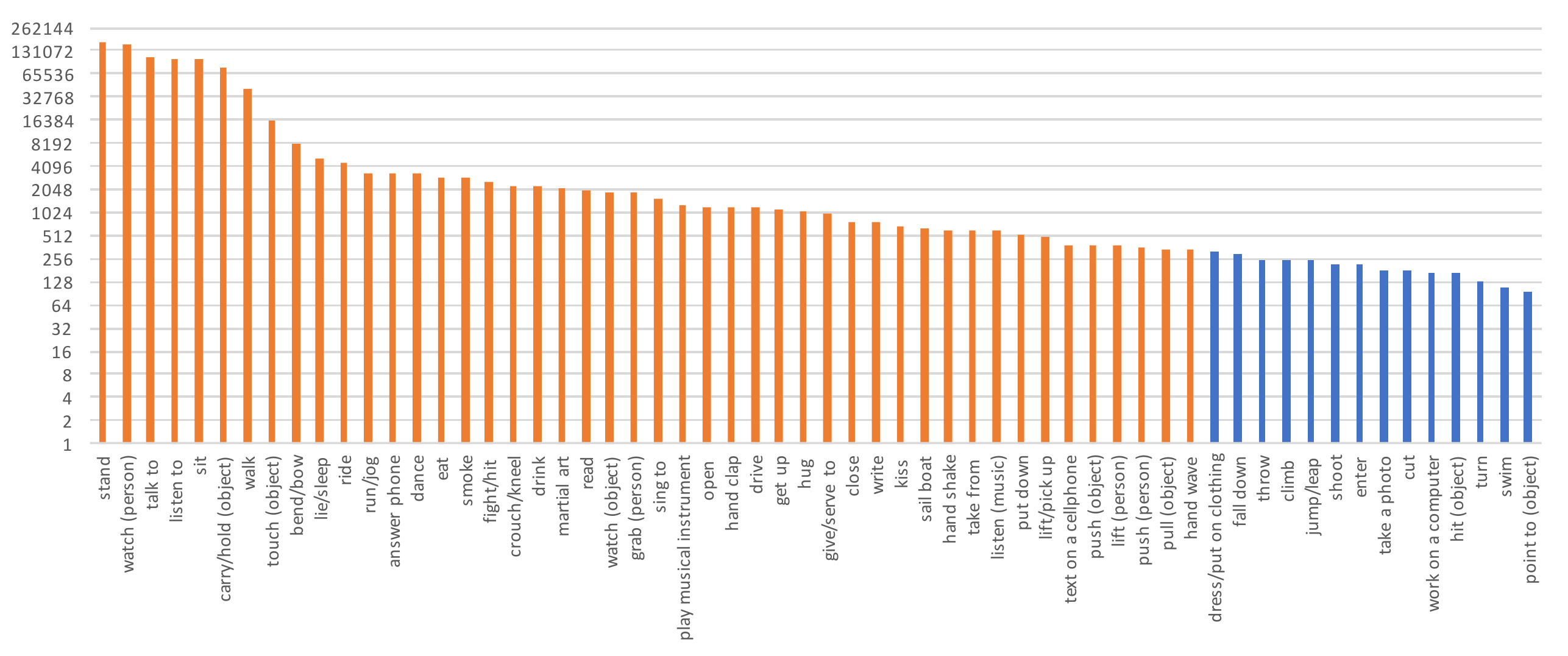}
\end{center}
   \caption{The number of training samples in head and tail classes used in the validation stage are shown in orange and blue respectively.  Notice that the plot is in logarithmic scale.}
\label{fig:head and tail}
\end{figure*}

This section proposes a new performance measure for detection in imbalanced datasets. 
We begin by discussing mAP - the standard measure for detection, and demonstrating that it is not well suited for measuring the performance on the tail categories. We then propose an alternative metric - mean sampled average precision (mSAP), which addresses the limitations of mAP in a principled way. Finally, we present an empirical analysis of the metrics, demonstrating that mSAP preserves the mAP's scores for the categories in the head, while providing more informative scores for the categories in the tail. In addition, we compare to one more alternative metric, the receiver operator characteristic (ROC)~\cite{provost1998case}, and show that is does not address the limitations of mAP as effectively as the proposed mSAP.

\subsection{The AP metric in imbalanced sets}
\label{sec:ap}

Consider a category in the AVA dataset, such as {\tt stand}. For any given model, the average precision for this category is computed as a mean of precision values over different recall levels, where precision is defined in Eq~\eqref{eq:-1}, and recall in Eq~\eqref{eq:0}.
\begin{align}
Precision &= \frac{TP}{TP + FP} \label{eq:-1},\\
Recall &= \frac{TP}{TP + FN} \label{eq:0}.
\end{align}
In these equations, $TP$ stands for true positives - the number of ground truth instances of {\tt stand} correctly predicted by the model, $FP$ for false positives - the number of instances other categories classified as {\tt stand}, and $FN$ is the number of false negatives - misclassified examples of {\tt stand}. This measure was originally introduced for information retrieval~\cite{manning2010introduction} to capture how accurate a model is in finding the instances of a certain class in a given collection. It was later adapted by the computer vision community for object~\cite{everingham2010pascal}, and subsequently, action detection~\cite{klaser2010human}. 

The test sets in these datasets were, however, artificially balanced, thus an information retrieval metric was suitable for comparing the performance of detection models on different categories. Indeed, if the proportion of two categories in a set is comparable, then the complexity of retrieving examples of these categories from the set is mainly determined by the ability of the model to recognize them. This, however, is not true for a highly imbalanced dataset, such as AVA. For instance, among the 93994 examples in the validation set, 44449 belongs to the category {\tt watch (a person)} and only 32 are examples of {\tt point to (a person)}. Suppose we have a model that gives truly random prediction at each recall level. According to the precision function in Equation ~\eqref{eq:-1}, the proportion of the true positive sample over all positive samples is equal to the ratio of positive samples in the test set. Therefore the AP scores of this random model for {\tt watch} and {\tt point to} are 0.473 and 0.0003 respectively. Clearly the model's capacity on two classes is supposed to be similar. However, we are not able to make this conclusion from the values of AP score. We argue that the main reason is that the precision in Equation~\eqref{eq:-1} is strongly influenced by the size of the pool of positive and negative examples for the category, and not only by the recognition performance of the model.

This observation demonstrates that information retrieval metrics are not suitable for measuring recognition performance in imbalanced datasets. We now propose an improved metric which is explicitly separating the complexity of action recognition from that of retrieving extremely rare instances from large data collections.


\begin{table*}[]
\centering
\begin{tabular}{|c|l|c|c|c|c|c|c|}
\hline
\multicolumn{2}{|c|}{\multirow{2}{*}{Actions}}                                              & \multicolumn{2}{c|}{SAP} & \multicolumn{2}{c|}{AP} &\multicolumn{2}{c|}{ROC-AUC} \\ \cline{3-8} 
\multicolumn{2}{|c|}{}                                                                      & Baseline     & Ours     & Baseline      & Ours   & Baseline      & Ours  \\ \hline
\multirow{5}{*}{\begin{tabular}[c]{@{}c@{}}Head \\ class\end{tabular}} & stand          & 78.1          & 77.9     & 75.6         & 76.0    & 78.9 &   79.2  \\
                                                                       & watch (a person) & 61.6          & 61.9   & 59.0         & 59.3    & 64.7 &   63.9  \\
                                                                       & talk to          & 70.4          & 70.0    & 65.2         & 64.7    &  73.9 &   75.2 \\
                                                                       & listen to       & 52.7          & 54.2    & 50.2         & 51.3   &  68.0 &   69.2   \\
                                                                       & sit             & 76.1          & 76.6    & 72.6         & 73.4   & 82.6 & 82.8     \\ \hline
\multirow{5}{*}{\begin{tabular}[c]{@{}c@{}}Tail\\ class\end{tabular}}  & work on computer    & 63.5          & 75.4   & 4.2          & 24.5    & 84.4    &  87.6 \\
                                                                       & hit (an object)    & 42.6          & 58.8    & 0.2          & 0.6    & 80.3 & 81.6   \\
                                                                       & trun (screwdriver)  & 50.5          & 52.3 &  0.2          & 0.3     &  80.9 &   81.2  \\
                                                                       & swim               & 80.8          & 84.1    & 57.4         & 74.6   & 89.4 & 86.1  \\
                                                                       & point to (object)     & 23.2          & 41.8   & 0.1          & 0.2  & 61.7& 73.2   \\ \hline
\multicolumn{2}{|c|}{Average} & 47.8          & 49.1   & 16.7          & 17.6  & 77.7& 77.9   \\ \hline

\end{tabular}
\vspace{4mm}
\caption{Comparison of our SAP metric to AP and ROC-AUC. Results for the baseline and our method are presented for the five head classes with the largest number of samples and the five tail classes with the smallest number of samples. The last row gives the average scores over all classes.}
\label{table:metric_example}
\end{table*}

\subsection{A better metric: sampled AP}

In order to minimize the influence of class imbalance in testing, while still relying on a retrieval-based metric, we propose to construct an independent and balanced retrieval problem for each class. To this end, we randomly sample a subset of negative boxes for a category to obtain a balanced pool of positive and negative examples, and then compute the standard AP on this set. We repeat this process multiple times and average the results to obtain the final sampled AP (SAP) score.

More formally, assume that $D$ is the test set and $X_c$ is the set of test samples that belong to class $c$. For each trial $i$, we randomly sample a set of negatives $\hat{X}_{ci}$  (examples from other categories than $c$), with $|\hat{X}_{ci}|=|X_c|$ from the set of all negative samples $D \setminus X_c$. We then use these two sets to compute the average precision $AP(X_c \cup \hat{X}_{ci})$. This process is repeated $N$ times and the final sampled AP of a class $c$ is computed as follows:
\begin{equation}
    SAP(c) = \frac{1}{N} \sum_{i=1}^{N} AP(X_c \cup \hat{X}_{ci}).
\label{func:subsample}
\end{equation}
Finally, we average the SAP scores for all the classes to obtain the mean sampled average precision (mSAP):
\begin{equation}
    mSAP = \frac{1}{K} \sum_{c \in C} SAP(c),
\end{equation}
where $K$ is the number of classes in the dataset.

This metric has several desired properties. First of all, it directly addresses the issue with the AP discussed in Section~\ref{sec:ap}: balancing the number of positive and negative examples in each sample enables the AP score to reflect the recognition capabilities of the model, and not the complexity of retrieving a few instances from a large set of distractors. Secondly, as we will show in the next section, it preserves the original AP scores for the categories in the head of the distribution. Finally, despite the fact that only a subset of the test set is used in each sample, the model is still being evaluated on the whole test set, given random sampling and enough trials .

\begin{figure}
\begin{center}
\includegraphics[width=1.0\linewidth]{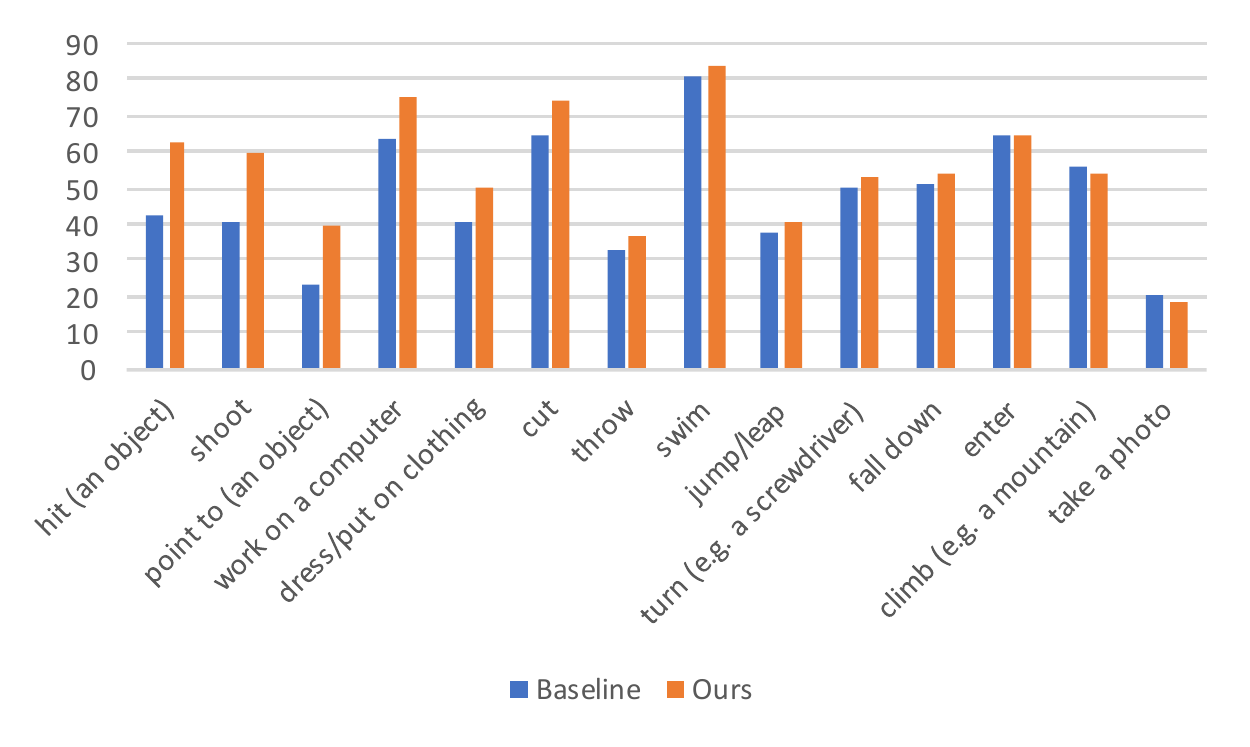}
\end{center}
\vspace{-5mm}
   \caption{The SAP scores of the baseline and our method for tail classes on the validation set. Results are ordered based on increase in performance between the two methods (left to right).}
\label{fig:tail_classes}

\end{figure}

\begin{table*}[]
\centering
\begin{tabular}{|l|c|c|c|}
\hline
\multirow{1}{*}{Methods}            
                         & \multicolumn{1}{c|}{All classes} & \multicolumn{1}{c|}{Tail classes} & Head classes \\ \hline
Baseline                 & \multicolumn{1}{c|}{47.8}        & \multicolumn{1}{c|}{47.7}         & 47.8                \\
Balancing \cite{Ling:1998:DMD:3000292.3000304}               & \multicolumn{1}{c|}{41.9}            & \multicolumn{1}{c|}{47.6}             &      40.2           \\
Focal loss \cite{Lin2017FocalLF}               & \multicolumn{1}{c|}{43.4}            & \multicolumn{1}{c|}{41.1}             &      44.2           \\
Ours                     & \multicolumn{1}{c|}{49.1}        & \multicolumn{1}{c|}{53.9}         & 47.7           \\ \hline
\end{tabular}
\vspace{4mm}
\caption{Comparison of our approach to state-of-the art balancing methods and the baseline. We report results with mean sampled AP for all classes, tail classes and head classes.}

\label{table:our training strategy}
\end{table*}

\begin{table}
\begin{center}
\begin{tabular}{|l|c|c|c|}
\hline
\multirow{1}{*}{Methods} 
 & \multicolumn{1}{c|}{All classes} & \multicolumn{1}{c|}{Tail classes} & Head classes\\ \hline
Baseline & \multicolumn{1}{c|}{47.8} & \multicolumn{1}{c|}{47.7} & 47.8  \\
1st stage all & \multicolumn{1}{c|}{47.4} & \multicolumn{1}{c|}{51.8} & 46.1  \\
1st stage head & \multicolumn{1}{c|}{49.1} & \multicolumn{1}{c|}{53.9} & 47.7 \\ \hline
\end{tabular}
\end{center}
\caption{\textbf{First stage training:} The mean sampled AP on validation set of both tail classes and head classes suggest that training the first stage with only head categories is better than with all categories. }
\label{table:stage one}
\end{table}

\subsection{Analysis}

We now experimentally analyze our proposed metric by comparing the performance of two models on the tail categories of AVA. One model is the baseline from~\cite{zhang2018structured}, whereas the other is our model that transfers the representation from head to tail categories (discussed in detail in the next Section). As can be seen from Table~\ref{table:metric_example}, the standard AP is capturing the difference between the two variants for the categories in the head, but the performance difference of the tail categories is in most cases close to 0, with the exception of {\tt swim} and {\tt work on computer}. These categories have very distinctive appearances, and are thus easy to separate from the others. This experimentally confirms our claim that AP is not a suitable metric for action detection in highly imbalanced datasets. In contrast, our SAP metric is not biased by the distribution of the test examples and provides informative results for all the categories in the tail. At the same time, it preserves the original AP scores for the head categories like {\tt stand} and {\tt watch}, since for these categories the random samples $X_c \cup \hat{X}_{ci}$ are close to the original distribution. Using the SAP metric we can discover that our method improves the recognition performance for the rare {\tt point to} category with respect to the baseline, whereas the standard AP score did not allow to make any conclusions.


Next, we compare our proposed metric to the area under the curve of ROC (ROC-AUC) in Table~\ref{table:metric_example}. 
The ROC curve  considers jointly the ratio of correctly classified positive samples (true positives) and the proportion of wrongly classified negative samples (false positives). For a class with a few positive samples, the false positive rate will be very low and thus the ROC will be high.  As shown in Table~\ref{table:metric_example}, the ROC scores are artificially high and do not reflect the performance differences for different methods on the tail classes.
In contrast, our proposed SAP is independent of the number of examples in the test set, as it uses sampling to balance the positives and negatives,
which makes it independent of the class distribution.


We also study the robustness of our proposed metric to the number of samples used in Eq.~\ref{func:subsample}. Figure \ref{fig:metric_sample_number} shows the mean and variance of the scores of the baseline model from~\cite{zhang2018structured} varying the value of $N$. We observe that the metric is relatively stable using as few as 15 samples, allowing to compute it efficiently even on datasets with a large number of categories. We also visualize the per-class SAP score and standard deviation of sampling 15 times for the baseline model in Figure \ref{fig:std_mean_metric} where the classes are sorted by the number of samples in decreasing order. Overall, the values of the standard deviation of both head classes and tail classes are small which proves that our proposed SAP metric is informative as well as practical.


\begin{table*}
\begin{center}
\begin{tabular}{|l|c|c|c|}
\hline
\multirow{1}{*}{Models}
 & All classes & Tail classes & Head classes \\ \hline
Baseline & 47.8 & 47.7 & 47.8  \\
2nd stage all & 47.3 & 50.9 & 46.2  \\
2nd stage classifier & 49.1 & 53.9 & 47.7  \\ \hline
\end{tabular}
\end{center}
\caption{\textbf{Second stage training:} Fine tuning only the final linear classifier weights to help transferring learned knowledge from head to tail categories consistently improves validation mSAP.}
\label{table:stage two}
\end{table*}

\vspace{-4mm}

\begin{table}
\begin{center}
\begin{tabular}{|l|c|c|c|}
\hline
\multirow{1}{*}{Models}
 & All& Tail& Head\\ \hline
Baseline & 47.8 & 47.7 & 47.8 \\
Ours + original distribution & 41.6 & 35.1 & 43.5 \\
Ours + balanced distribution & 49.1 & 53.9 & 47.6 \\ \hline
\end{tabular}
\end{center}
\caption{\textbf{Second stage training:} We show the importance of balancing the dataset in the second stage of training by comparing the resulting mSAP of training with the balanced training set and the original training set.}
\label{table:balance}
\end{table}

\section{Learning in the tail}
Having developed an informative metric for evaluating a model's performance in the tail, we now focus on studying different strategies for knowledge transfer from the data-rich head to the rare tail categories. We begin with standard approach of balancing during training~\cite{Ling:1998:DMD:3000292.3000304,Lin2017FocalLF}, see Section~\ref{sec:simple}. We observe that in the action detection domain this leads to  poor performance. To address this issue, we propose an alternative approach, which learns a features representation 
on the head classes, see Section~\ref{sec:linear_transfer}. Finally, in Section~\ref{sec:ablation} 
we examine different variants of our approach. 

\subsection{State-of-the-art approaches to balancing}
\label{sec:simple}

Balancing the training set is a simple strategy that is widely used to handle the long-tail distribution problem. Following the design principle of \cite{Ling:1998:DMD:3000292.3000304},  we balance the training set by duplicating samples of the tail classes so that each class has roughly the same number of samples. We directly train the action detection model with the balanced set. As shown in Table \ref{table:our training strategy}, we obtain the mAP score of 11.3 compared to the original baseline with 16.7. Specifically, we observe that the performance of the tail classes remains low, whereas the performance of the head classes actually drops. We argue that this decrease is the result of the fact that the model fails to learn a useful representation from this highly redundant set, remembering that in AVA, the categories in the tail are thousands times less frequent than the head categories. Therefore, in the balanced training set the examples of rare categories are oversampled, and most of the batches are not informative. Instead we propose to explicitly split the training set into head and tail parts and use the former to learn a generic representation for action detection. We then transfer this representation to learn to detect rare categories, experimenting with several transfer strategies.

Next, we compare to another balancing approach proposed in object detection. Lin et al.~\cite{Lin2017FocalLF} designed the focal loss to down-weight the gradients of well-classified examples, and thus facilitate learning of rare categories. We show that the performance of this approach is limited in our case.
We train our baseline with focal loss, setting the focusing parameter $\lambda$ to 2 (the optimal value according to~\cite{Lin2017FocalLF}), and report the results in Table~\ref{table:our training strategy}.
First, observe that the focal loss results in a significant performance decrease on AVA. This is because the focal loss was designed for relatively balanced datasets, such as COCO. In contrast, our method can successfully transfer the representation from head to tail categories, even when head categories have a thousand times more examples.

\subsection{Our training approach for action detection}
\label{sec:linear_transfer}

The problem of identifying a subset of categories in a dataset that are useful for learning a transferable representation is not trivial. We first observe that the selected classes should be both representative of the whole set of categories and have enough examples to learn a rich feature representation. Therefore, we want to select all the categories for which a model is able to learn a strong representation. In other words, the categories that are defined as tail classes should be those classes which have too few samples for model to learn. We use the difference on the AP score of a baseline model on the training set and test sets as our main criterion for selecting the classes in the head. The intuition is that for the classes which have too few training samples the training AP scores will be even lower than test AP scores, indicating that the model is not able to adapt to them. The split is shown in Figure \ref{fig:head and tail} with head classes marked with orange color and tail classes in blue color.




Given the split into head and tail classes, we first train our model only on the information-rich head categories. Next, in the second stage, we freeze the model, and learn a linear classifier using the whole dataset with balancing, where balancing is done is exactly the same way as in the baseline described in Section~\ref{sec:linear_transfer}. The whole process is shown in the Figure \ref{fig:model}. The overall training schema is elaborated in Algorithm \ref{alg:one}. We show our mean sampled AP and the standard mAP of the action detection baseline on all action classes, tail classes and head classes respectively in Table ~\ref{table:our training strategy}. As we can see in the table, our proposed method achieves a significant increase of 6.2\% mSAP on the tail classes. We compare  the SAP performance of our proposed method and the baseline approach on the tail classes in Figure \ref{fig:tail_classes}. We observe a significant increase on the classes like "work on a computer" and "cut" which have a discriminative appearance, but were hard for the baseline to learn due to the imbalance in the training set. In contrast, our proposed transfer learning schema allows to better recognize these classes even from a few examples.

\begin{algorithm}
   \caption{Our Proposed Training Schema}
    \begin{algorithmic}[]
    \State \textbf{Input:} A set of head classes samples $\mathcal{H}$, a set of tail classes samples $\mathcal{T}$
    \State \textbf{Output:} A model $\mathcal{F}$ consisting of an I3D backbone and a linear classifier
    \State Train a set of weights $\theta^*$ for I3D, and $\hat{w}$ for linear classifier with SGD to minimize the loss $L$ on sample ($x, y$) of head classes.
    \begin{equation*}
        \{ \theta^*, \hat{w} \} = \operatorname*{argmin}_{\theta, w} \mathop{\mathbb{E}}_{(x,y)\sim \mathcal{H}} [L(\mathcal{F}_{\theta, w}(x), y)]
    \end{equation*}
    \State Fix $\theta^*$, train for a set of weights $w^*$ for linear classifier with SGD to minimize the loss $L$ on balanced dataset.
    \begin{equation*}
        w^* = \operatorname*{argmin}_{w} \mathop{\mathbb{E}}_{(x,y)\sim Balanced(\mathcal{H}, \mathcal{T})} [L(\mathcal{F}_{\theta^*, w}(x), y)]
    \end{equation*}
    \State $\theta^*$ and $w^*$ are the final weights we use for action detection.
\end{algorithmic}
\label{alg:one}
\end{algorithm}




\subsection{Ablation of variants of our approach}
\label{sec:ablation}

We now analyze the effectiveness of our approach of excluding the tail classes from the first stage of training in Table~\ref{table:stage one}. To this end we explore the alternative strategy of training the model on all the categories without balancing and then re-learning a linear classifier with balancing. This variant still shows an improvement on the tail categories over the baselines due to balancing in the second stage, but fails to preserve the performance of head categories. This is due to the fact that the tail classes have very few training samples, on which it is hard for the model to learn generalizable features. Our proposed training schema, on the other hand, shows an even larger performance boost on the tail classes and at the same time, allows the model to have similar performance on head classes which shows that we can maximally exploit the model's capacity by training I3D feature representation only with head classes.


Next, we study different ways of transferring representation from head to tail classes in Table \ref{table:stage two}. In particular, we analyze two approaches: finetuning the whole model, and only learning a linear classifier on top of a fixed representation. As can be seen from the table, both approaches results in an increase in performance on the tail classes, though improvement of our approach is larger. With our mean sampled AP metric, our method achieves an improvement of 3\% on tail classes. Note that another good feature of fixing the feature representation during the transfer stage is that it also helps preserve the performance on head classes. We hypothesize that this is due to the fact that updating the I3D on a highly redundant, balanced training set results in it loosing some of its generalization abilities.

Finally, we evaluate how balancing the training data during the second stage influences the performance. Table \ref{table:balance} compares the results on the balanced training set as well as the original training set respectively under our proposed two-stage training schema. As we can see from the table, without balancing the model is biases towards head classes with far more samples, resulting in a significant decrease of  18\% mSAP on the tail classes. 



\section{Conclusion}

In this paper we studied action detection in the real world, where the goal is to learn a model that is able to recognize both the head classes with hundreds of thousands of samples, as well as the tail classes with only a few samples. Different from the other works, we first analyzed the data imbalance problem in the test set. We demonstrated that the standard mAP metric is not suitable for measuring the performance of the tail classes. We then proposed a new, more informative measure - mean sampled average precision (mSAP), which takes balanced samples from the test set and averages the sample scores. 

We then used the new measure to study the problem of class imbalance in the training set. We proposed a simple training schema where features are learnt for the head classes and are then transferred to the tail. We also show that training a simple linear classifier with balancing in the second stage on top of fixed representation is both an efficient and effective transfer strategy. Our proposed method can be used to boost the performance of existing action detection models in a simple way.

{\small
\bibliographystyle{ieee}
\bibliography{egbib}
}


\end{document}